%% file: flip_v1.tex
\documentclass[cameraready]{Interspeech}

\usepackage{amsmath}
\usepackage{booktabs}
\usepackage{comment}
\usepackage[table]{xcolor}
\usepackage{xcolor}
\usepackage{realboxes}
\usepackage{graphicx}
\graphicspath{{figures/}}
\usepackage{amssymb,amsfonts}
\usepackage{multicol}
\usepackage{multirow}
\usepackage{textcomp}
\usepackage{footmisc}
\usepackage{pifont}
\usepackage{xcolor}
\usepackage{tikz}
\usetikzlibrary{bayesnet}
\usetikzlibrary{arrows}
\input{math_def.tex}

\hypersetup{
    colorlinks,
    linkcolor={black},
    citecolor={black},
    urlcolor={black}
}

\title{FLiP: Towards understanding and interpreting multimodal multilingual sentence embeddings}

\author[orcid=0000-0002-3725-742X]{Santosh}{Kesiraju}
\author[orcid=0000-0001-9852-3456]{Bolaji}{Yusuf}
\author[orcid=0009-0005-9875-3012]{\v{S}imon}{Sedl\'{a}\v{c}ek}
\author[orcid=0000-0001-7938-3945]{Old\v{r}ich}{Plchot}
\author[orcid=0000-0002-2281-9637]{Petr}{Schwarz}

\address{
    Speech@FIT, Brno University of Technology, Czechia
}

\email{\{kesiraju,iyusuf,isedlacek,iplchot,schwarzp\}@fit.vut.cz}

\newcommand{\cmark}{\ding{51}}
\newcommand{\xmark}{\ding{55}}

\usepackage{xcolor}


\begin{document}

\maketitle

\begin{abstract}

This paper presents factorized linear projection (FLiP) models for understanding pretrained sentence embedding spaces. We train FLiP models to recover the lexical content from multilingual (LaBSE), multimodal (SONAR) and API-based (Gemini) sentence embedding spaces in several high- and mid-resource languages. We show that FLiP can recall more than 75\% of lexical content from the embeddings, significantly outperforming existing non-factorized baselines. Using this as a diagnostic tool, we uncover the modality and language biases across the selected sentence encoders and provide practitioners with intrinsic insights about the encoders without relying on conventional downstream evaluation tasks. Our implementation is public\footnote{ \texttt{\url{https://github.com/BUTSpeechFIT/FLiP}}}.
\end{abstract}

\section{Introduction}
\label{sec:intro}
Learning semantically aligned sentence representations agnostic of the underlying language or modality (speech, text) has several applications ranging from retrieval~\cite{reimers-gurevych-2019-sentence}, classification~\cite{mdhaffar-2025-sense} and building parallel datasets across language pairs~\cite{duquenne-2021-mining,duquenne-etal-2023-speechmatrix,heffernan-etal-2022-bitext,pothula25_interspeech}. Despite the limitations due to the \textit{compression} into a \textit{single vector}, the research and development of embeddings models (or sentence encoders) is evolving and spanning across modalities such as image-text~\cite{radford-etal-2021-clip}, and audio-text~\cite{clap-2023}. Interpreting such single vector embeddings has several applications including (a) getting a sense of why a certain embedding has been retrieved (b) as a diagnostic tool for researchers and engineers to build better models, (c) as a scientific pursuit to know ``\textit{what can you cram into a single {\$}{\&}!{\#} vector?}''~\cite{conneau-etal-2018-cram}.

Interpreting embeddings or, in general, the parameters of deep neural networks has gained significant attention in recent years. Among these, the linear representation hypothesis~\cite{park:2023:LRH} is one of the simplest, elegant and theoretically grounded frameworks because of the use of linear (interpretable) models (or projections or operations) for interpreting black-box models or embeddings. While most research and findings focus on studying text-only~\cite{adi:2017}, image-text~\cite{Bhalla:2024:Splice} or audio-text~\cite{Zhang:2025:Transformation} embeddings and often rely on English language for interpretation, little attention has been devoted to a comprehensive analysis of the embedding space across languages and modalities (speech and text). 

Log-linear models have been the core of embedding based models such as word2vec~\cite{mikolov-wor2vec}, paragraph vector~\cite{mikolov-doc2vec} and their variants~\cite{Kesiraju:2016:SMM}, and have shown to scale well with data. We build on the concepts from prior works and apply them for interpreting semantically-aligned sentence embeddings that are supposed to be agnostic to modality and the underlying language. Under consistent experimental settings, we also compare the ``linearity of the semantics" across embeddings models such as SONAR~\cite{sonar:2023}, LaBSE~\cite{feng-etal-2022-language} and a black-box (API-based) Gemini embedding~\cite{lee-etal-2025-gemini-embedding}. Although we make comparisons, a clear distinction has to be made here: our methodology should be primarily seen as diagnosis tool and not as a replacement for massive benchmarks like MTEB~\cite{muennighoff-etal-2023-mteb} that rank models based on various downstream task abilities. Instead our comparisons are complementary to MTEB as we dive deeper into the representation abilities of the embeddings (thereby the underlying encoders) and quantify them with more than a single number.



\begin{enumerate}
    \item We show that a \textit{well trained} factorized linear projection (FLiP is sufficient to recall ~75-80\% of the lexical content from a \textit{well encoded} sentence embeddings, demonstrating that semantic concepts are linearly represented in embedding space.
   \item We use FLiP as a diagnostic tool to systematically analyze modality alignment, language alignment, and concept language effects across SONAR, LaBSE, and Gemini embedding spaces.
   \item We compare FLiP with SpLiCE~\cite{Bhalla:2024:Splice} and show that it is objectively a better tool for interpreting the embeddings.
\end{enumerate}

\section{Methodology}
\label{sec:method}
We formulate the task of interpreting embeddings via a proxy task of keyword (concept) extraction via simple linear projection (LiP). 
First, we introduce LiP model as general framework and then present the proposed factorized variant and the cross-lingual and cross-modal training schemes.


\subsection{Log-linear model}
\label{sec:vanilla}
The vanilla form of the model learns a single projection matrix ($\mb{W}$) and a bias vector ($\mb{b}$) that maps a sentence embedding to the space of the vocabulary. Formally, let $\mb{t}_n \in \mathbb{R}^d$ be a sentence embedding extracted from a text using a pre-trained model such as SONAR. Let $\mb{x}_n \in \mathbb{N}_{0}^{|\mathcal{V}|}$ be the bag-of-words vector of the underlying text, where each element indicates the absolute count of the number of occurrences for word indexed at position $i$ in an ordered vocabulary set $\mathcal{V}$. For each sentence embeddings, we compute the probability distribution over the words $\boldsymbol{\theta}_n \in \triangle^{|\mathcal{V}|-1}$ i.e.,\ the simplex of dimension $|\mathcal{V}|-1$,  using the aforementioned linear projection, followed by $\mathrm{softmax}$.
%
\begin{equation}
\label{eq:mono_theta}
 \bs{\theta}_n = \mathrm{softmax}(\mb{b} + \mb{W} \mb{t}_n)   
\end{equation}
 The model is optimized to maximize the regularized log-likelihood of the training data of $N$ samples.
 \begin{equation}
     \mathcal{L} = \sum_{n=1}^{N} \Big[ \mb{x}_n^{\T} \, \log \bs{\theta}_n \Big] - \mathcal{R}(\mb{W})
 \end{equation}
where $\mathcal{R}(\mb{W})$ denotes regularization over model parameters (word embedding matrix). 

\subsection{Cross-modal and cross-lingual training}
\label{sec:crossmodal}
Since existing sentence encoders are trained to yield semantically aligned sentence embeddings for speech and text across languages, we introduce the cross-modal or cross-lingual signal into the training objective. This allows our model to compensate for any variability in language or modality (speech, text) present in the embeddings. Let $\mb{s}_n \in \mathbb{R}^d$ be the speech utterance-level embedding corresponding to textual counterpart $\mathbf{t}_n$. Projecting speech embedding on to \textit{same} $\mathbf{W}$ and normalizing gives:
\begin{equation}
\label{eq:speech_phi}
\bs{\phi}_n = \mathrm{softmax}(\mb{b} + \mb{W}\mb{s}_n)
\end{equation}
Combining this with \eqref{eq:mono_theta} results in the following objective
\begin{equation}
\label{eq:joint_objective}
\mathcal{L} = \sum_{n=1}^N  \alpha\mb{x}_n{\T} \log(\bs{\theta}_n) + (1-\alpha)\mb{x}_n^{\T} \log(\bs{\phi}_n) - \mathcal{R}(\mb{W})
\end{equation}
where $\alpha$ balances text and speech contributions.

The above formulation~\eqref{eq:joint_objective} applies to bilingual text-text scenario, where $\mathbf{t}_n$ and $\mathbf{s}_n$ can be seen as text embedding pairs from primary (the one matching the language of the vocabulary) and secondary languages respectively. 

\subsection{Factorized log-linear model}
\label{sec:factorization}
While the vanilla model only trains the single matrix $\mathbf{W}$, we find it beneficial to perform factorization (potentially low-rank) of this matrix during training: $\mathbf{W}=\mathbf{AB}$, 
%
where $\mb{A} \in \mathbb{R}^{|\mathcal{V}| \times r}$ is the word embedding matrix and $\mb{B} \in \mathbb{R}^{r \times d}$ is the modality-to-latent projection. Rank $r \leq d$ is shared across text and speech.

Even though at full-rank ($r=d$) such factorization does not change the expressive power of the transformation (it remains purely linear), in practice, we observe superior experimental results compared to the non-factorized model. We recognize that this is mainly due to the implicit regularization mechanism that the factorization imposes on $\mathbf{W}$~\cite{suriya-2017-fact}. Additionally, making the factorization of $\mathbf{W}$ low-rank makes the training more efficient (in terms of memory requirements) due to the vocabulary dimension being substantially smaller compared to the original sentence embedding dimension, while still reaping the benefits of the implicit regularization, as is demonstrated in Section~\ref{sec:res_rank}.

\noindent \textbf{Regularization.}
We incorporate $L_1$ regularization on $\mb{A}$ to induce sparsity. We use proximal gradient descent with a soft-thresholding operator to get explicit zeros in our solution. 

\subsection{FLiP: Keyword extraction from embeddings}
\label{sec:inference}
Keyword extraction requires a simple linear projection. Given an unseen sentence embedding from text or speech $\mb{u} \in \mathbb{R}^{d}$, we compute the logits $\mb{z} = \mb{b} + \mb{A} (\mb{B}\,\mb{u})$
and select the top-$k$ indices of $\mathbf{z}$, which we map to their corresponding words in the vocabulary. This process is efficient and does not require any additional optimization or search at inference time. We note that the bias vector $\mathbf{b}$ typically learns a $\log$-prior distribution over the vocabulary. This gives us a leverage to compensate the scores over frequent (stop words) words as will be shown in Section~\ref{sec:res_ner}.

\section{Experimental setup}
\label{sec:setup}

\subsection{Datasets and languages}
\label{sec:datasets}
For cross-modal experiments, we used speech-text pairs for English, German, and French from the Mozilla Common Voice (MCV) corpus~\cite{CV:2020} (v15.0). The standard splits comprises of 1.7M speech, text pairs for English, 0.5M pairs each for German (DE) and French (FR). The corresponding dev and test set had roughly 16k speech, text pairs per language. 

For cross-lingual experiments, we used parallel text data for two pairs from Europarl~\cite{koehn-2005-europarl} (English--\{German, French\}) and four pairs from Samanantar~\cite{ramesh-etal-2022-samanantar} (English--\{Bengali (BN), Hindi (HI), Tamil (TA), Telugu (TE)\}). For each language pair , we constructed training sets with roughly 1.8M bitexts and 10k bitexts each for dev and test sets. 

All the text is lower cased and stripped off punctuation. The vocabulary size is fixed at 100K words (unigrams) per language unless otherwise explicitly stated.

\subsection{Sentence embedding models}
\label{sec:embedding_models}


SONAR~\cite{sonar:2023} is a high-coverage, modality-agnostic encoder supporting 200 written and 37 spoken languages. It functions by first establishing a shared multilingual text space using a Transformer encoder, then mapping speech utterances into that same space via knowledge distillation. This creates a unified 1024-dimensional embedding where semantically equivalent text and speech are geometrically aligned.

LaBSE~\cite{feng-etal-2022-language} is a multilingual text-only model covering 109 languages, built on a dual-encoder BERT architecture. It is trained on massive datasets (17 billion monolingual sentences and 6 billion translation pairs) using masked language modeling and translation ranking tasks. The model produces 768-dimensional $L_2$-normalized embeddings, ensuring that cross-lingual translations reside near each other in the vector space.

Gemini Embeddings~\cite{lee-etal-2025-gemini-embedding} represent a ``black-box" API-based approach, utilizing a bi-directional model initialized from a decoder-only LLM. It generates text-only embeddings via mean pooling of the output sequence. While it defaults to 3072 dimensions, it supports flexible output sizes; this specific study utilizes the 768-dimensional variant for storage efficiency and performance balance.



\subsection{Evaluation}
\label{sec:evaluation}
We employ multiple evaluation metrics to assess the quality of the extracted keywords. We lowercase the text and remove punctuations before evaluation.

\noindent \textbf{Accuracy.} We set $k$ to the number of in-vocabulary reference tokens for each individual sentence embedding and compute the fraction of top-$k$ extracted keywords that match these reference tokens.

\noindent \textbf{Span-aware accuracy.} To account for multi-word concepts ($n \ge 2$), we extract in-vocabulary reference spans via greedy left-to-right matching. The longest matching $n$-gram at each position is selected as a single reference unit before advancing.

\noindent \textbf{Jaccard index.} To assess cross-lingual and cross-modal alignment, we compute the Jaccard index between the correctly extracted keyword sets (hit sets) of two independent models on the same test utterances. This measures inter-model consistency, revealing if different models activate the same linear semantic concepts.

\noindent \textbf{Named entity recall$@k$.} A subset of our test data is automatically tagged with named entities using Gemini-2.5-Flash-Lite to evaluate recall over the top-$k$ extracted words. For multi-word entities, we report \textbf{strict} recall (requiring all constituent unigrams) and \textbf{partial} recall (crediting any retrieved constituent token).

%

\subsection{Baselines}
\label{sec:baselines}
We compare FLiP against two baselines. The first is the full-rank LiP (without factorization), which serves as a proxy to prior linear probing works~\cite{conneau-etal-2018-cram,park:2023:LRH}. 

\noindent \textbf{SpLiCE}~\cite{Bhalla:2024:Splice}. Our second baseline decomposes multimodal embeddings into sparse, interpretable concepts. Following the original methodology for constructing a concept vocabulary from training transcripts, we filter unigrams and bigrams based on corpus frequency ($f_{\mathrm{min}}=20$) and pointwise mutual information ($\mathrm{PMI}_{\mathrm{min}}=1.5$) of the bigrams as estimated from the training corpus. From the filtered candidates, we select the top terms by frequency to form a final vocabulary of $|\mathcal{V}|=10000$ concepts (6500 unigrams and 3500 bigrams).

\subsection{Proposed models}
\label{sec:proposed_models}

We train FLiP and the baseline LiP under identical settings for a fair comparison. Models are optimized using AdamW~\cite{ilya-2019-adamw} ($\eta = 5\mathrm{e}{-3}$, halved when the evaluation metric plateaus) for up to 100 iterations with a batch size of 6000. Early stopping is based on the unigram recall on the development set.

We tune hyperparameters based on development set performance, sweeping the factorization rank $r \in \{128, \ldots, 1024\}$, $L_1$ penalty on $\mb{A}$ ($\lambda_1 \in \{0, \ldots 1\mathrm{e}{-2}\}$), and $L_2$ weight decay ($\lambda_2 \in \{0, \ldots 1\mathrm{e}{-2}\}$). Following the formulations in Section~\ref{sec:crossmodal}, our default cross-modal and cross-lingual experiments use a balanced mixing coefficient ($\alpha=0.5$), though we also train independent models with $\alpha \in \{0.0, 1.0\}$ to isolate specific modalities or target languages.

\section{Results and analysis}
\label{sec:results}

\begin{table}[!t]
\centering
\caption{Mean accuracy ($\pm$ std.err) of keyword extraction across different ranks for English text and speech SONAR embeddings from Mozilla Common Voice.}
\resizebox{0.92\columnwidth}{!}{
\begin{tabular}{cccrr}
\toprule
\textbf{Factorized} & \textbf{Dimension} & \textbf{Text} & \textbf{Speech} \\
\midrule
\xmark & Full & 59.45 ($\pm$ 0.13) & 57.27 ($\pm$ 0.13) \\
\cmark & Full & \textbf{77.29} ($\pm$ 0.12) & \textbf{74.09} ($\pm$ 0.14) \\
\cmark & 512 & 76.77 ($\pm$ 0.12) & 73.62 ($\pm$ 0.14) \\
\cmark & 256 & 74.39 ($\pm$ 0.14) & 71.67 ($\pm$ 0.14) \\
\cmark & 128 & 67.48 ($\pm$ 0.13) & 65.81 ($\pm$ 0.14) \\
\bottomrule
\end{tabular}
}
\vspace{-0.3cm}
\label{tab:rank_analysis}
\end{table}

\subsection{Factorization and rank analysis}
\label{sec:res_rank}

In Table~\ref{tab:rank_analysis}, we compare the keyword extraction performance of different training configurations of LiP -- namely the factorization and/or rank of the FLiP model. All models are trained on both speech and text SONAR embeddings of Common Voice English (we set $\alpha=0.5$) with an English vocabulary. We observe that factorizing the vocabulary matrix is crucial to achieving best performance likely due to the implicit regularization imposed on the matrix in this scenario. Even at lower ranks, FLiP outperforms the vanilla LiP model by a significant margin, with the 512-dimensional model (76.77\% accuracy on text) showing just a marginal performance degradation compared to the full 1024-dimensional FLiP while having only half the parameters. For subsequent experiments, all presented FLiP models are trained in this 512-dimensional setting. We observed the $L_1$ and $L_2$ regularization had marginal effect on the results and we set $\lambda_1=1e-4$ and $\lambda_2=0$ for subsequent experiments.



%

\subsection{Modality alignment}
\label{sec:res_alignment}

%
 Here, we probe cross-modal linear semantic structure in SONAR by comparing keywords extracted via FLiP trained on either speech-only or text-only ($\alpha$ is set to 0 or 1). The models are trained on EN, DE, and FR independently, with each language constituting both the text and speech training data as well as the vocabulary. Apart from standard accuracy, we also report Jaccard index between independent models against the reference model. As seen in Table~\ref{tab:cross_modal_sonar}, when confined within a single language, SONAR shows good cross-modal alignment between speech and text embeddings, as the keyword extraction performance is comparable when training on speech or text, where training on speech brings slightly better transferability to the textual domain.

\begin{table}[!t]
\centering
\caption{Probing results for cross-modal (speech, text) linear semantic alignment. Results in shaded cells are the absolute differences in accuracy with the adjacent number.}
\resizebox{0.9\columnwidth}{!}{
\begin{tabular}{lcrrc}
\toprule 
& &  \multicolumn{3}{c}{\textbf{Test embeddings}}  \\
  &   \textbf{Training} &  \multicolumn{2}{c}{{Accuracy}}& {{Jaccard Index}} \\
\textbf{Lang.} & \textbf{embeddings} &  {Text}  & {Speech}  & (Text, Speech) \\ 
\midrule
\multirow{2}{*}{EN} & Text     & 75.71   &   \cellcolor{lightgray} --2.22 & 87.20 \\
& Speech   &  \cellcolor{lightgray} --0.14  &  72.68  & 89.58 \\ 
\midrule 
\multirow{2}{*}{DE}  & Text & 60.11  &  \cellcolor{lightgray} --2.43  & 81.90 \\ 
             & Speech & \cellcolor{lightgray}  --1.93   &  60.60      &  84.22  \\
\midrule 
\multirow{2}{*}{FR} & Text &  58.48  &  \cellcolor{lightgray} --3.42  & 78.60 \\
& Speech & \cellcolor{lightgray} --1.61   &  58.98      &  83.01     \\
\bottomrule
\end{tabular}  
}\vspace{-0.2cm}
\label{tab:cross_modal_sonar}
\end{table}

\subsection{Language alignment}
\label{sec:res_concept_lang}

Next, we test whether the SONAR embedding space is semantically linear across languages. For this task, we chose the paired text datasets described in Section~\ref{sec:datasets}. The vocabulary is fixed to EN, and the models are trained always with EN as the baseline and subsequently on one of the six paired languages (DE, FR, BN, HI, TA, TE). The results are shown in Table~\ref{tab:cross_lingual_sonar}.

The primary observation is that for SONAR, English is consistently the preferred language for linear concept representation, where training on EN will yield good accuracy on linguistically similar languages like DE or FR. However, the other direction always results in a significant performance gap for the non-EN language compared to EN, possibly due to the model failing to learn better linear decomposition of the non-EN embedding space, which is slightly degraded compared to EN. When the languages become more linguistically dissimilar (BN, HI, TA, TE), however, training on EN no longer brings the same performance transfer, suggesting that the representation spaces of such languages are not well aligned compared to EN.

\begin{table}[!ht]
\centering
\caption{Probing results for cross-lingual semantic alignment in the embedding space. Results in shaded cells are the absolute differences in accuracy with the adjacent number.}
\resizebox{0.75\columnwidth}{!}{
\begin{tabular}{crrc}
\toprule 
&  \multicolumn{3}{c}{\textbf{Test embeddings}}  \\
\textbf{Training} &  \multicolumn{2}{c}{{Accuracy}}& {Jaccard Index} \\
\textbf{embeddings}     &  {EN}  & {XX}  & ({EN},\, {XX}) \\ 
\midrule
 EN       &  70.81   & \cellcolor{lightgray} --5.55 &  80.79    \\
 DE       &  \cellcolor{lightgray} --2.22 &  54.76   &  77.10   \\ 
\midrule 
 EN      &  70.38  & \cellcolor{lightgray} --5.58  &  80.75    \\
 FR      & \cellcolor{lightgray} --2.22 &   53.27  &  76.43  \\ 
\midrule 
 EN     &  75.17  & \cellcolor{lightgray} \textbf{--24.05} &  55.44      \\
 BN     & \cellcolor{lightgray} --3.1  &  53.91   &  74.66      \\ 
\midrule 
 EN     &  67.69  & \cellcolor{lightgray} {--15.07}  &  62.11    \\
 HI     & \cellcolor{lightgray} --3.49  & 56.08    &  64.49  \\ 
 \midrule 
EN     &  70.98  &  \cellcolor{lightgray} \textbf{--30.87} &  46.97     \\
TA     & --3.28 \cellcolor{lightgray}  & 47.51  &  70.65     \\ 
\midrule 
EN       &  85.62  &  \cellcolor{lightgray} \textbf{--36.24} & 50.49     \\
TE     & \cellcolor{lightgray} --9.48 & 62.99  &  71.20    \\ 
\bottomrule
\end{tabular}    
}
\label{tab:cross_lingual_sonar}
\end{table}

\subsection{Language of the vocabulary}
To ablate the effect of using the the target/non-target vocabulary for cross-lingual scenarios, we retrain the models from the previous Section on both languages ($\alpha=0.5$), once having the vocabulary in EN and once in the other language in the pair. Again, Table~\ref{tab:vocab_change} the proclivity of SONAR for representing concepts in EN is significantly highlighted here, being especially pronounced for the more dissimilar language pairs like EN-TA, which yield poor performance for both languages when the vocabulary language is not set to EN.

\begin{table}[!ht]
\centering
\caption{Probing results for cross-vocabulary relations. Keyword extraction accuracy for bilingual-trained models with EN and XX vocabulary, tested on EN and XX embeddings. XX is the secondary language other than EN from each row appropriately.}
\scalebox{0.85}{
\begin{tabular}{ccc|cl}
\toprule
\textbf{Training} &  \multicolumn{2}{c|}{{\textbf{Vocabulary (EN)}}}& \multicolumn{2}{c}{{\textbf{Vocabulary (XX)}}} \\
{\textbf{Pair}}     &  {\textbf{EN-emb}}  & {\textbf{XX-emb}} ($\Delta$)  & {\textbf{XX-emb}} & {\textbf{EN-emb}} ($\Delta$) \\
\midrule
EN-DE  & 69.44  &  54.99 (--14.45) &  54.14 &  44.17 (--9.97) \\
EN-FR  & 69.04  &  53.87 (--15.17) &  54.27 &  45.03 (--9.24) \\
EN-BN  & 73.79  &  54.09 (--19.70) &  36.99 &  26.57 (--10.42) \\
EN-HI  & 66.92  &  56.65 (--10.27) &  55.45 &  46.18 (--9.27)  \\
EN-TA  & 69.00  &  48.70 (--20.30) &  19.46 &  14.26 (--5.20)   \\
EN-TE  & 83.38  &  62.66 (--20.72) &  44.36 &  29.30 (--15.06)  \\
\bottomrule
\end{tabular}
}
\label{tab:vocab_change}
\end{table}

\subsection{Comparison across embedding models}
\label{sec:res_models}


We additionally evaluate FLiP when trained on embeddings extracted from SONAR, LaBSE~\cite{feng-etal-2022-language}, and Gemini embedding~\cite{lee-etal-2025-gemini-embedding}. Again, the Europarl EN-DE dataset is used, and two FLiPs are trained for each embedding model; one with the EN vocabulary and one with DE. The results shown in Table~\ref{tab:comparion_encoders} once again demonstrate the tendency of embedding models to better align with the English vocabulary, with SONAR offering best keyword extraction performance across the board.

\begin{table}[!ht]
    \centering
    \caption{Keyword extraction accuracy of FLiP models trained on  sentence embeddings from various encoders.}
\resizebox{0.75\columnwidth}{!}{
    \begin{tabular}{crr|rr}
    \toprule
    & \multicolumn{2}{c}{\textbf{Vocabulary (EN)}} & \multicolumn{2}{c}{\textbf{Vocabulary (DE)}} \\
     \textbf{Encoder}    &  \textbf{EN}   & \textbf{DE} & \textbf{DE} & \textbf{EN}\\
     \midrule
      SONAR     & \textbf{69.44} & \textbf{54.99} & 54.14 & 44.17 \\
      LaBSE     & 60.22 & 50.19 & 49.16 & 40.72 \\
      Gemini    & 60.94 & 49.60 & 47.78 & 38.83 \\
      \bottomrule
    \end{tabular}
    }
    \label{tab:comparion_encoders}
\end{table}

\subsection{Named entity recall}
\label{sec:res_ner}
\begin{figure}
    \centering
    \includegraphics[width=\linewidth]{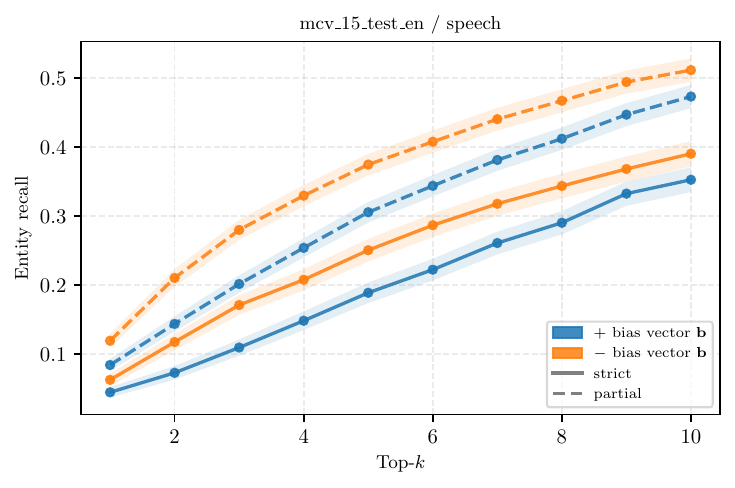}
    \caption{Named-entity recall as function of top-$k$ extracted keywords and the effect of bias vector on the logit scores.}
    \label{fig:ner}
\end{figure}

Fig.~\ref{fig:ner} shows strict and partial entity recall as function of top-$k$ evaluated on MCV (EN) speech embeddings. Recall increases monotonically with $k$, and interestingly, removing the bias term improves entity recall -- the bias acts as a log unigram prior that up-weights frequent stop words, which compete with and displace named entities in the top-$k$. Consider the sentence ``\textit{Eisen was born in Brooklyn New York and was raised on Staten Island}." At $k=5$, the model extracts \textit{brooklyn, born, raised, was, in} -- recovering \textit{Brooklyn} but missing \textit{New York} and \textit{Staten Island}. At $k=10$ it retrieves \textit{brooklyn, york, new, staten, island}, achieving strict recall of 1.0 across all three entities. 



\subsection{Comparison with SpLiCE}
\label{sec:splice_comparison}
For a fair comparison with SpLiCE, we used identical vocabulary of 10000 concept vocabulary and trained FLiP on Mozilla common voice English using cross-modal training. Using SpLiCE we decomposed each of the SONAR text, speech embedding into its constituent concepts. We evaluate using span-aware accuracy and the results presented in Table~\ref{tab:baseline_splice} show that the FLiP is able to extract almost twice the number of keywords from the embeddings as compared to SpLiCE.

\begin{table}[!ht]
\centering
\caption{Span-aware accuracy ($\pm$se) comparison of FLiP with SpLiCE on Mozilla CV (EN) SONAR embeddings.}
\resizebox{0.7\columnwidth}{!}{
\begin{tabular}{lrr}
\toprule
\textbf{Method} & \textbf{Text} & \textbf{Speech} \\ 
\midrule
SpLiCE & 29.58 ($\pm$0.19) & 28.21 ($\pm$0.19) \\
FLiP & \textbf{61.45} ($\pm$0.20) & \textbf{58.83} ($\pm$0.21) \\
\bottomrule
\end{tabular}
}
\label{tab:baseline_splice}
\end{table}

\section{Conclusions}
\label{sec:conclusion}



This paper introduced FLiP, a factorized log-linear model for interpreting multimodal and multilingual sentence embeddings. Framing interpretation as a linear keyword extraction task, we showed that well-aligned embedding spaces linearly encode most of their lexical content, recalling over 75\% of vocabulary concepts via a single projection. Furthermore, FLiP proved significantly more effective and parameter-efficient at keyword extraction than existing baselines like SpLiCE, at the same time not requiring any heuristics for vocabulary selection. Probing SONAR, LaBSE, and Gemini revealed that while intra-language cross-modal alignment is robust, cross-lingual representations remain heavily English-biased, with semantic linearity degrading in linguistically distant languages.

\section{Generative AI Use Disclosure}
Generative AI tools were used at the skeleton level of the paper to explore various options for organizing the structure (sections, and subsections), and transposing existing Tables to alternative perspective quickly. Seldom used to paraphrase a few lines.

\section{Acknowledgments}
The work was supported by Ministry of Education, Youth and Sports of the Czech Republic (MoE) through the OP JAK project ``Linguistics, Artificial Intelligence and Language and Speech Technologies: from Research to Applications'' (ID:CZ.02.01.01/00/23\_020/0008518), and by European Union’s Horizon Europe project No. SEP-210943216 ``ELOQUENCE'' . Computing on IT4I supercomputer was supported by MoE through the e-INFRA CZ (ID:90254).

\bibliographystyle{IEEEtran}
\bibliography{mybib}

\end{document}

%% file: math_def.tex
\newcommand{\mb}[1]{\mathbf{#1}}
\newcommand{\bs}[1]{\bm{#1}}

\newcommand*{\T}{^{\mkern-1.1mu\mathsf{T}}}     

\newcommand{\myfrac}[2]{%
  \setbox0\hbox{$#1$}        
  \dimen0=\wd0               
  \setbox1\hbox{$#2$}        
  \dimen1=\wd1               
  \ifdim\wd0<\wd1            
  \dfrac{\hfill#1}{#2}     
  \else
  \dfrac{#1}{\hfill#2}     
  \fi
}
